# BIOMETRIC VERIFICATION OF HUMANS BY MEANS OF HAND GEOMETRY[1]

**Marcos Faundez-Zanuy**
**Dep. de Telecomunicaciones y Arquitectura de computadores**
**Escola Universitària Politècnica de Mataró, Adscrita a la UPC**
**Avda. Puig i Cadafalch 101-111, 08303 Mataró (Barcelona), SPAIN**
**e-mail: faundez@eupmt.es**

## ABSTRACT

This Paper describes a hand geometry biometric identification system. We have acquired a database of 22 people, 10 acquisitions per person, using a conventional document scanner. We propose a feature extraction and classifier. The experimental results reveal a maximum identification rate equal to 93.64%, and a minimum value of the Detection Cost Function equal to 2.92% using a Multi Layer Perceptron Classifier.

## 1. INTRODUCTION

In recent years, hand geometry has become a very popular biometric access control, which has captured almost a quarter of the physical access control market [1]. Even if the fingerprint [2-4] is the most popular access system, the study of other biometric systems is interesting, because the vulnerability of a biometric system [5] can be improved using some kind of data fusion [6] between different biometric traits. This is a key point in order to popularize biometric systems [7], in addition to privacy issues [8].

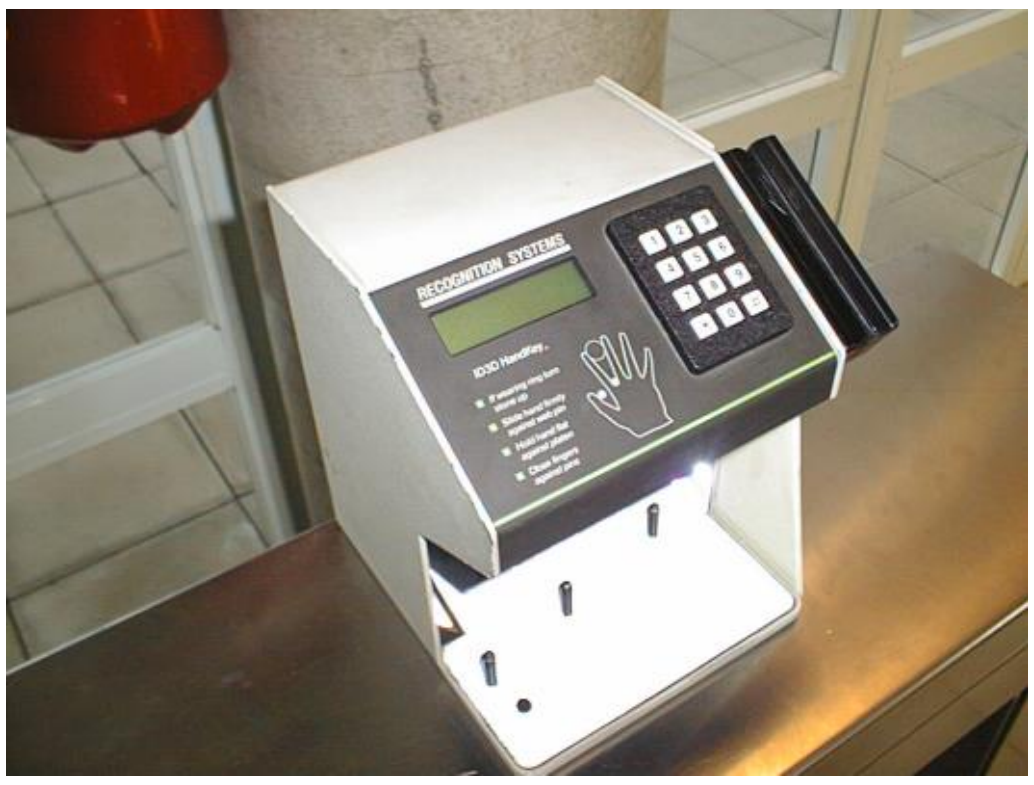

*Figure. 1. Commercial three-dimensional scanner.*

Although some commercial systems, such us the system shown in figure 1 rely on a three-dimensional profile of the hand, in this paper we study a system based on two dimensional profiles. Although three dimensional devices provide more information than two dimensional ones, they require a more expensive and voluminous hardware.

A two-dimensional profile of a hand can be get using a simple document scanner, which can be purchased for less than 100 USD. Another possibility is the use of a digital camera, whose cost is being dramatically reduced in the last years.
In our system, we have decided to use a conventional scanner instead of a digital photo camera, because it is easier to operate, and cheaper. On the other hand, although a digital camera is extremely fast in taking a photo, the last generation scanners (such as EPSON 4870 Photo perfection) are able to capture a DIN A4 size colour document (24 bit) at a 150 dpi resolution in less than 15 seconds when using the USB 2 port, which is a quite reasonable time.
This paper can be summarized in three main parts: section two describes a database which has been specially acquired for this work. In section three, we describe the pre-processing and we study the discrimination capability of several measurements on the sensed data. Section four provides experimental results on identification and verification rates using neural net classifiers.

## 2. DATABASE

We have acquired a database of 22 people, and 10 different acquisitions per person. If some acquisition has not let to extract some of the parameters described in the next section, this capture has been rejected and replaced by a new one. Figure 2 shows an example of defective acquisitions and the reason.

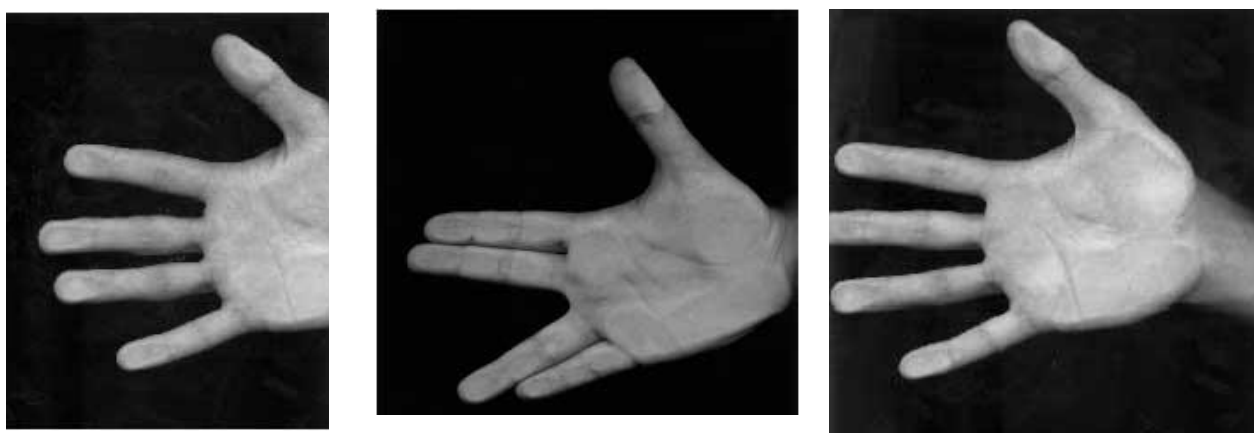

***Figure 2.** Example of defective acquisitions. The first one is defective because it is cut on the base. In the second one, some fingers are joined. In the third one, one finger is cut.*

The database has been stored in bmp format using 8 bits per pixel (256 gray levels), a resolution of 100 dpi, and an image size of 216x240 mm. Higher resolutions would imply more details but also more computational time in

[1] This work has been supported by FEDER and the Spanish grant MCYT TIC2003-08382-C05-02

M. Faundez-Zanuy, "Biometric verification of humans by means of hand geometry," Proceedings 39th Annual 2005 International Carnahan Conference on Security Technology, 2005, pp. 61-67, doi: 10.1109/CCST.2005.1594816.

order to process a hand image. In our preliminary experiments we have found that 100 dpi offers a good compromise. Obviously this resolution is insufficient for other related applications such as palm print, which is analogous to fingerprint recognition, but using the ridge and valley pattern of the hand skin. Thus, the system will rely on the silhouette of the hand and will ignore other details such as fingerprints, lines, scars and colour. For this reason, the first step of the pre-processing described in the next section will be a binarization and a contour extraction. Although this procedure discards useful information for discrimination, it also alleviates other problems, such as the perspiration of the skin which blots the thin details of the image. Figure 3 shows an example of this phenomenon.

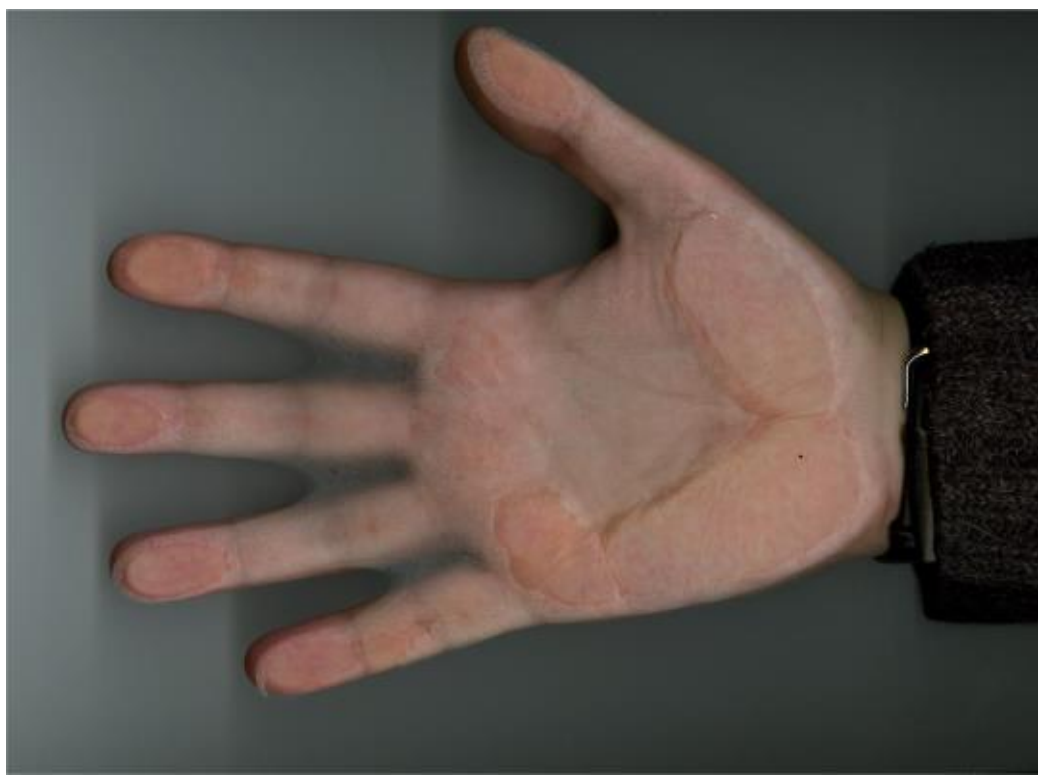

***Figure 3.*** *Example of hand acquisition at 150 dpi and 24 bit per pixel (color image), with perspiration problem. This problem can be neglected after the binarization step.*

## 3. FEATURE EXTRACTION

Some measurements are extracted from each hand image. This section describes the proposed algorithm for feature extraction.

### Preprocessing algorithm

Figure 4 shows a block diagram of the pre-processing algorithm.

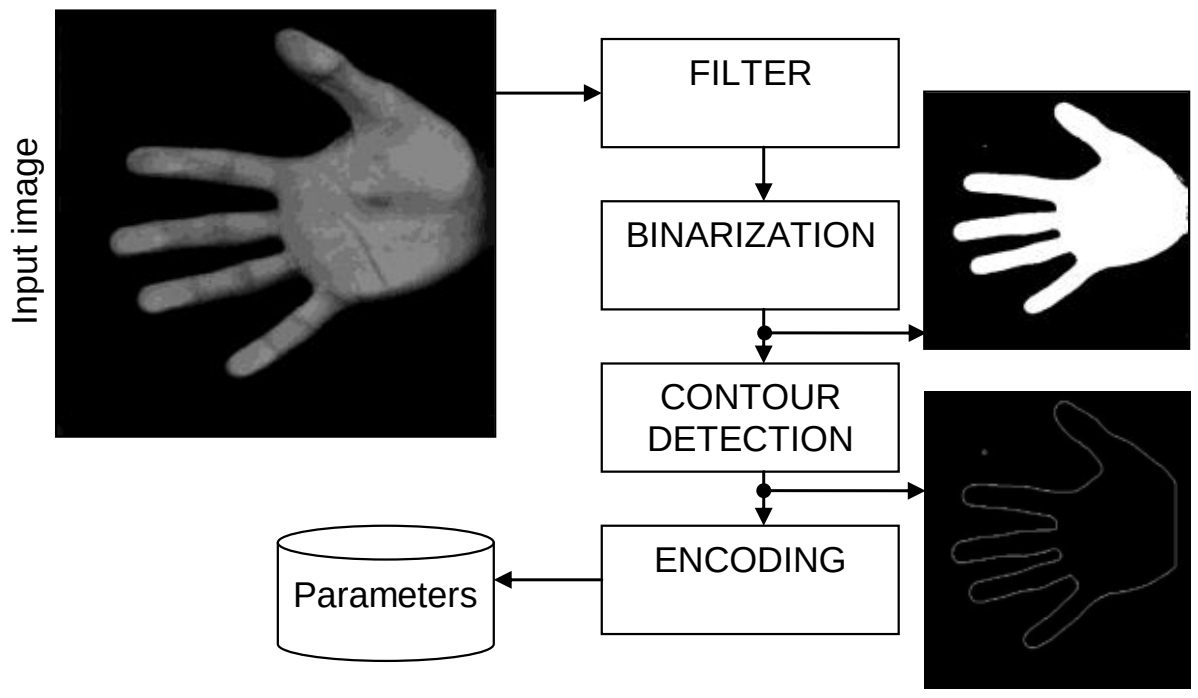

*Figure 4. Block diagram for the proposed pre-processing scheme.*

The description of each block is the following:

### Filter

We apply a low-pass filtering in order to remove spurious noise.

### Binarization

The goal is the conversion from an image $I(x,y)$ at 8 bit per pixel to a monochrome image $I'(x,y)$ (1 bit per pixel. "0"=black, "1"=white), applying a threshold:

$$I'(x,y)=\begin{cases}1 & if\ I(x,y)\geq threshold\\ 0 & otherwise\end{cases} \quad (1)$$

We use threshold=0.07, which was experimentally obtained.

### Contour detection

The goal is to find the limits between the hand and the background. For this purpose the algorithm detects the intensity changes, and marks a closed set of one pixel wide and length the perimeter of the hand. Edge points can be thought of as pixel locations of abrupt grey-level change. For example it can be defined an edge point in binary images as black pixels with at least one white nearest neighbor. We use the Laplacian of Gaussian method, which finds edges by looking for zero crossings after filtering the image with a Laplacian of Gaussian filter.

### Coding

This step reduces the amount of information. We translate a bmp file to a text file that contains the contour description. The encoding algorithm consists of a chain code. In chain coding the direction vectors between successive boundary pixels are encoded. Figure 5 shows our code, which employs 8 possible directions and can be coded with 3-bit codewords.

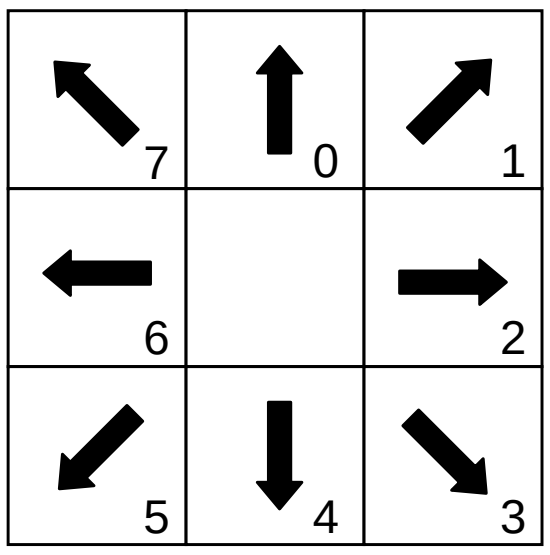

***Figure 5.*** *Contour coding algorithm.*

Once upon the chain code is obtained, the perimeter can be easily computed: for each segment, an even code implies +1 and an odd code $+\sqrt{2}$ units. The beginnings and endings of fingers and wrist are found looking for minimum and maximum values in the chain code.

The finger limits (base and maximum height) are detected in the middle of a region with a "5" and "3"·code. Figure 6 shows, for example, the maximum of the first and middle fingers.

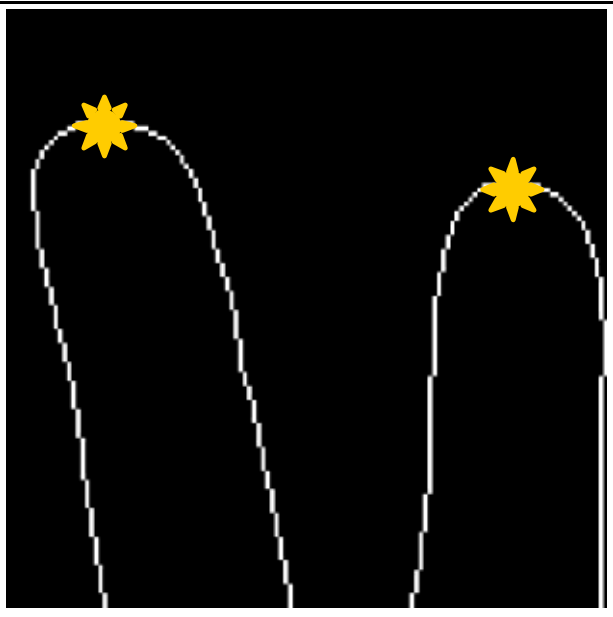

*Figure 6. Maximum of first and middle fingers.*

**Proposed features**

Using the result of the previous section as input, we propose the following measurements (see figure 7):

1. Thumb finger length.
2. First finger length.
3. Middle finger length.
4. Ring finger length.
5. Little finger length.
6. Wrist length.
7. Thumb base width.
8. First finger width.
9. Middle finger width.
10. Ring finger width.
11. Little finger width.
12. Hand perimeter.
13. Hand surface.

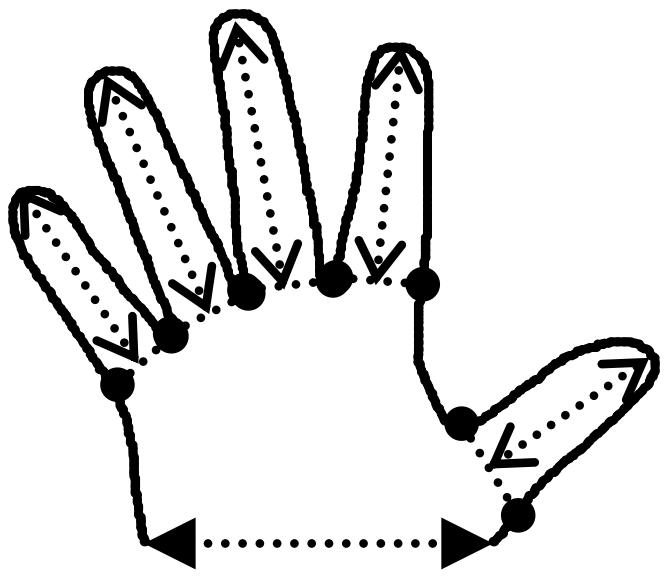

*Figure 7. Measured features.*

Some of these features have been removed in the experiments due to their low discrimination capability. Our experiments have revealed that results are improved deleting features 1, 6, 7 and 13. Thus, we will select the remaining nine features per image.

## 4. EXPERIMENTAL RESULTS

**Conditions of the experiments**

Our results have been obtained with the database described in section 2, the preprocessing of section 3, and the selected parameters of section 4, in the following situation: 22 persons, images 1 to 5 for training, and images 6 to 10 for testing.

**Identification**

Figure 8 shows the general scheme of a biometric system.

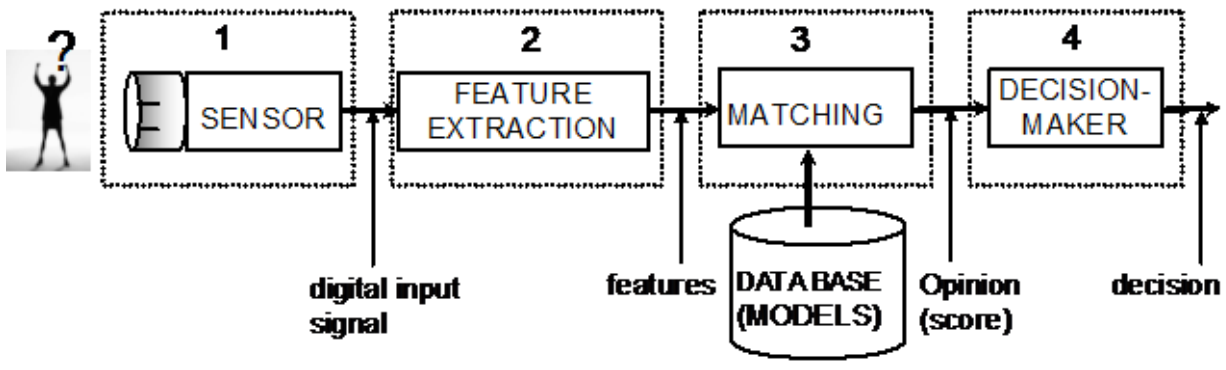

*Figure 8. General scheme of a biometric recognition system.*

These systems can be operated in two ways:

a) Identification: In this approach no identity is claimed from the person. The automatic system must determine who is trying to access.

b) Verification: In this approach the goal of the system is to determine whether the person is who he/she claims to be. This implies that the user must provide an identity and the system just accepts or rejects the users according to a successful or unsuccessful verification. Sometimes this operation mode is named authentication or detection.

For identification, if we have a population of $N$ different people, and a labelled test set, we fill up a matrix $S$, where the elements are interpreted in the following way:

$$s_{ijk} = O[j]\Big|_{\vec{x}\in person\#i} \quad ,k = 1,\cdots \#trials \qquad \textbf{(2)}$$

Where trials is the number of different testing images per person ($k$=5 in our experiments), and $s_{ijk}$ is the similarity from the $k$ realization of an input signal belonging to person $i$, to the model of person $j$.

This matrix can be drawn as a three dimensional data structure (see figure 9).

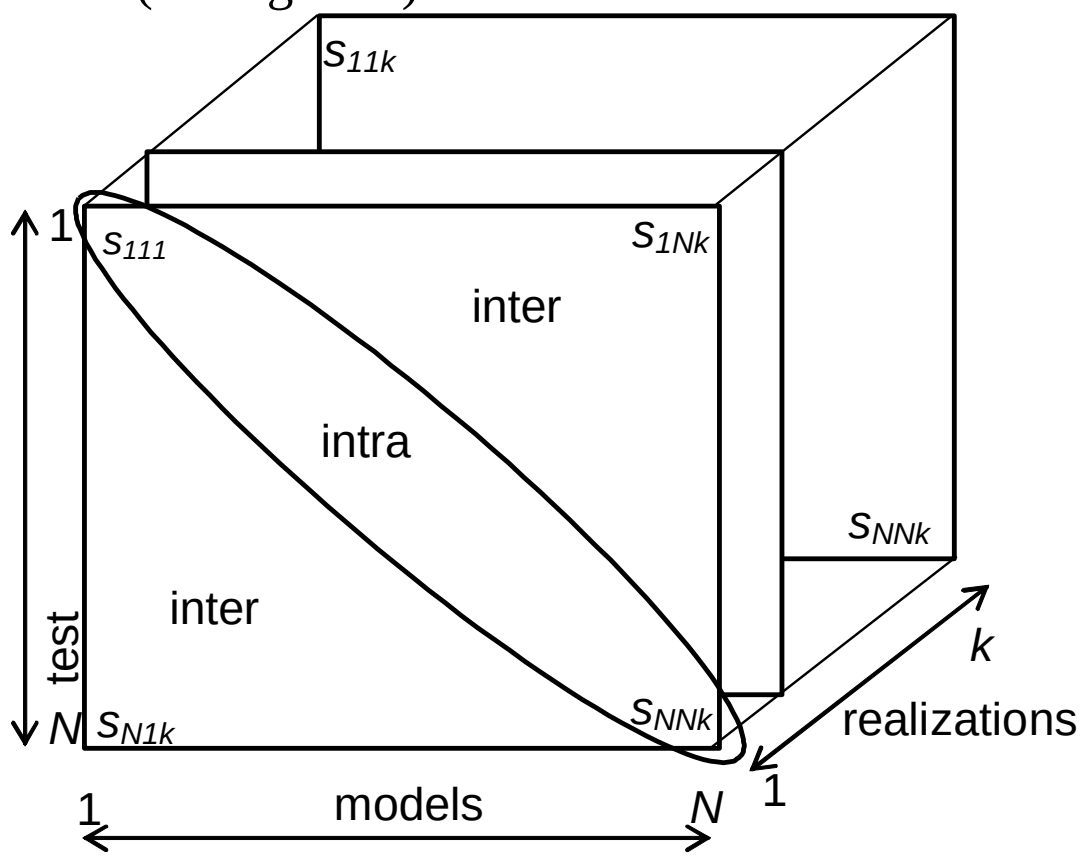

*Figure 9. Proposed data structure.*

Thus, the identification rate looks for each realization, in each raw, if the maximum similarity is inside the principal diagonal (success) or not (error), and works out the identification rate as the ration between successes and number of trials (successes + errors):

```
for i=1:N,
    for k=1:#trials,
             if(siik>sijk) ∀j≠i, then
               success=success+1
             else error=error+1
             end
    end
```

```
end
```

**Verification**

Verification systems can be evaluated using the False Acceptance Rate (FAR, those situations where an impostor is accepted) and the False Rejection Rate (FRR, those situations where a user is incorrectly rejected), also known in detection theory as False Alarm and Miss, respectively. There is trade-off between both errors, which has to be usually established by adjusting a decision threshold. The performance can be plotted in a ROC (Receiver Operator Characteristic) or in a DET (Detection error trade-off) plot [17]. DET curve gives uniform treatment to both types of error, and uses a logarithm scale for both axes, which spreads out the plot and better distinguishes different well performing systems and usually produces plots that are close to linear. DET plot uses a logarithmic scale that expands the extreme parts of the curve, which are the parts that give the most information about the system performance. For this reason the speech community prefers DET instead of ROC plots. Figure 10 shows an example of DET plot.

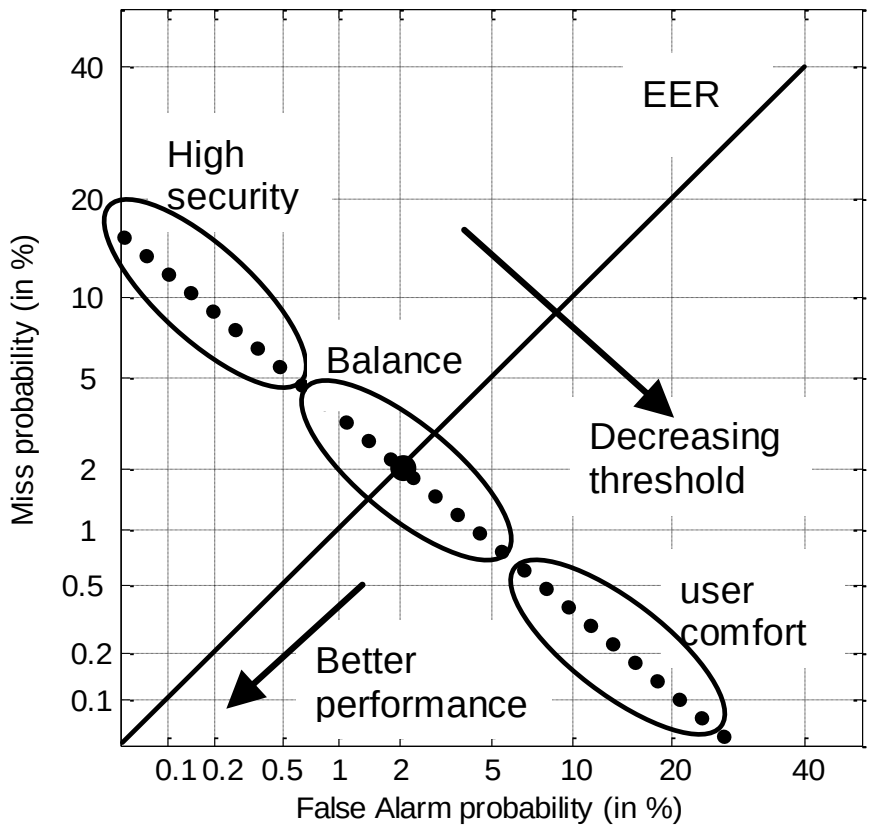

*Figure 10. Example of a DET plot for a user verification system (dotted line).*

We have used the minimum value of the Detection Cost Function (DCF) for comparison purposes. This parameter is defined as [17]:

$$DCF = C_{miss} \times P_{miss} \times P_{true} + C_{fa} \times P_{fa} \times P_{false} \quad (3)$$

Where $C_{miss}$ is the cost of a miss (rejection), $C_{fa}$ is the cost of a false alarm (acceptance), $P_{true}$ is the a priori probability of the target, and $P_{false} = 1 - P_{true}$ , $C_{miss} = C_{fa} = 1$.

Using the data structure defined in figure 9, we can easily apply the DET curve analysis. We just need to split the distances into two sets: intra-distances (those inside the principal diagonal), and inter-distances (those outside the principal diagonal).

**Nearest neighbor classifier**

We obtain one model from each training image. During testing each input image is compared against all the models inside the database (22x5=110 in our case) and the model close to the input image (by means of an error criterion) indicates the recognized person.

In our experiments we are using, for each user, all other users' samples as impostor test samples, so we finally have, $N$=22×5 (client)+22×21×5 (impostors)=2420 different tests. We have used two different distance measures:

$$MSE(\vec{x},\vec{y}) = \sum_{i=1}^{P} (x_i - y_i)^2 \quad (4)$$

$$MAD(\vec{x},\vec{y}) = \sum_{i=1}^{P} |x_i - y_i| \quad (5)$$

Where $P$ is the vector dimension.

Using the Nearest neighbor (NN) classifier, we get the results showed in table 1.

| Classifier | Identification rate | DCF |
|---|---|---|
| NN (MSE) | 64.55% | 14.44% |
| NN (MAD) | 73.64% | 11.41% |

*Table 1. Identification and verification rates for Nearest Neighbor classifier*

**Multi-Layer Perceptron classifier trained in a discriminative mode**

We have trained a Multi-Layer Perceptron (MLP) [9] as discriminative classifier in the following fashion: when the input data belongs to a genuine person, the output (target of the NNET) is fixed to 1. When the input is an impostor person, the output is fixed to –1. Figure 11 shows the neural network architecture. We have used a MLP with 30 neurons in the hidden layer, trained with the Levenberg-Marquardt algorithm, which computes the approximate Hessian matrix, because it is faster and achieves better results than the classical back-propagation algorithm. We have trained the neural network for 10 epochs (50 epochs when using regularization). We also apply a multi-start algorithm and select the best result.

The input signal has been fitted to a [–1, 1] range in each component.

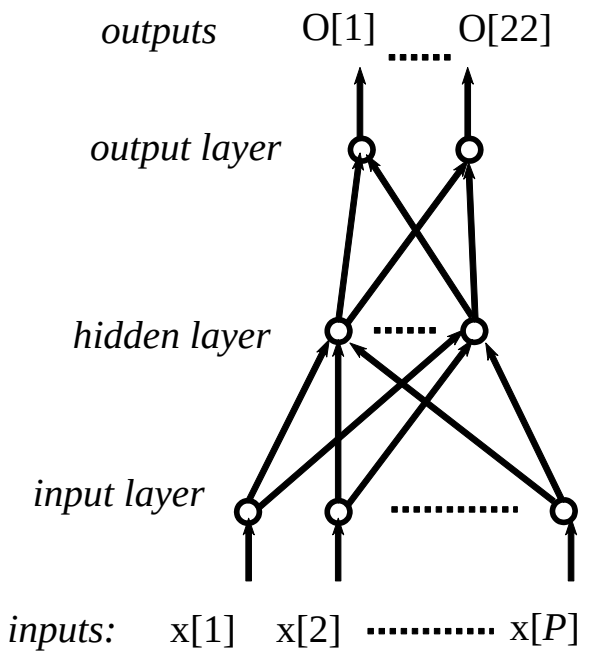

*Figure 11. Multi-Layer Perceptron architecture.*

One of the problems that occur during neural network training is called *overfitting*: the error on the training set is driven to a very small value, but when new data is presented to the network the error is large. The network has memorized the training examples, but it has not learned to generalize to new situations. The adopted solution to the overfitting problem has been the use of regularization. The regularization involves modifying the performance function, which is normally chosen to be the sum of squares of the network errors on the training set. So, this technique helps take the mystery out of how to pick the number of neurons in a network and consistently leads to good networks that are not overtrained.
The classical Mean Square Error (MSE) implies the computation of (6):

$$MSE = \frac{1}{N}\sum_{i=1}^{P}(t_i - a_i)^2 \qquad (6)$$

Where $t$, $a$ are the $P$ dimensional vectors of the test input and the model, respectively. The regularization uses the following measure (7):

$$MSEREG = \gamma MSE + (1-\gamma)\frac{1}{n}\sum_{j=1}^{n} w_j^2 \qquad (7)$$

Thus, it includes one term proportional to the modulus of the weights of the neural net.
In addition, there is another important topic: the random initialization. We have studied two strategies:

a) To pick up the best random initialization (the initialization which gives the higher identification rate)
b) A committee of neural networks, which combines the outputs of several MLP, each one trained with a different initialization [16].

**Error correction codes**

Error-control coding techniques [12] detect and possibly correct errors that occur when messages are transmitted in a digital communication system. To accomplish this, the encoder transmits not only the information symbols, but also one or more redundant symbols. The decoder uses the redundant symbols to detect and possibly correct whatever errors occurred during transmission.
Block coding is a special case of error-control coding. Block coding techniques map a fixed number of message symbols to a fixed number of code symbols. A block coder treats each block of data independently and is a memoryless device. The information to be encoded consists of a sequence of message symbols and the code that is produced consists of a sequence of codewords. Each block of $k$ message symbols is encoded into a codeword that consists of $n$ symbols; in this context, $k$ is called the message length, $n$ is called the codeword length, and the code is called an [$n$, $k$] code.
A message for an [$n$, $k$] BCH (Bose-Chaudhuri-Hocquenghem) code must be a $k$-column binary Galois array. The code that corresponds to that message is an $n$-column binary Galois array. Each row of these Galois arrays represents one word.
BCH codes use special values of $n$ and $k$:

- $n$, the codeword length, is an integer of the form $2^m-1$ for some integer $m > 2$.
- $k$, the message length, is a positive integer less than $n$.

However, only some positive integers less than $n$ are valid choices for $k$. This code can correct all combinations of $t$ or fewer errors, and the minimum distance between codes is:

$$d_{\min} \geq 2t+1 \qquad (8)$$

Table 2 shows some examples of suitable values for BCH codes.

| *n* | *k* | *t* |
|---|---|---|
| 7 | 4 | 1 |
| 15 | 11 | 1 |
| | 7 | 2 |
| | 5 | 3 |
| 31 | 26 | 1 |
| | 21 | 2 |
| | 16 | 3 |
| | 11 | 5 |
| | 6 | 7 |

Table 2. Examples of values for BCH codes.

**Multi-class learning problems via error-correction output codes.**

Multi-class learning problems involve finding a definition for an unknown function $f(\vec{x})$ whose range is a discrete set containing $k > 2$ values (i.e. $k$ classes), and $\vec{x}$ is the set of measurements that we want to classify. The definition is acquired by studying large collections of training examples of the form $\{\vec{x}_i, f(\vec{x}_i)\}$.
We must solve the problem of learning a $k$-ary classification function $f: \Re^n \to \{1,\cdots,k\}$ from examples of the form $\{\vec{x}_i, f(\vec{x}_i)\}$. The standard neural network approach to this problem is to construct a 3-layer feed-forward network with $k$ output units, where each output unit designates one of the $k$ classes. During training, the output units are clamped to 0.0, except for the unit corresponding to the desired class, which is clamped at 1.0. During classification, a new $\vec{x}$ value is assigned to the class whose output unit has the highest activation. This approach is called [13-14] the ***one-per-class*** approach, since one binary output function is learnt for each class.
An alternative method, proposed in [13-14] and called ***error-correcting output coding*** (ECOC), gives superior performance. In this approach, each class $i$ is assigned an $m$-bit binary string, $c_i$, called a codeword. The strings are chosen (by BCH coding methods) so that the Hamming distance between each pair of strings is guaranteed to be at least $d_{min}$. During training on example $\vec{x}$, the $m$ output units of a 3-layer network are clamped to the appropriate binary string $c_{f(\vec{x})}$. During classification, the new example $\vec{x}$ is assigned to the class $i$ whose codeword $c_i$ is closest (in Hamming distance) to the $m$-element vector of output activations. The advantage of this approach is that it can recover from any $t = \left\lfloor \frac{d_{\min} - 1}{2} \right\rfloor$ errors in learning the individual output units. Error-correcting codes act as ideal distributed representations.

In [13-14] some improvements using this strategy were obtained when dealing with some classification problems, such as vowel, letter, soybean, etc., classification. In this paper, we apply this same approach for biometric recognition based on hand-geometry measurements.

| *k* | code |
|---|---|
| 1 | 1,0,0,0,0,0,0,0,0,0,0,0,0,0,0,0,0,0,0,0,0,0 |
| 2 | 0,1,0,0,0,0,0,0,0,0,0,0,0,0,0,0,0,0,0,0,0,0 |
| 3 | 0,0,1,0,0,0,0,0,0,0,0,0,0,0,0,0,0,0,0,0,0,0 |
| 4 | 0,0,0,1,0,0,0,0,0,0,0,0,0,0,0,0,0,0,0,0,0,0 |
| 5 | 0,0,0,0,1,0,0,0,0,0,0,0,0,0,0,0,0,0,0,0,0,0 |
| 6 | 0,0,0,0,0,1,0,0,0,0,0,0,0,0,0,0,0,0,0,0,0,0 |
| 7 | 0,0,0,0,0,0,1,0,0,0,0,0,0,0,0,0,0,0,0,0,0,0 |
| 8 | 0,0,0,0,0,0,0,1,0,0,0,0,0,0,0,0,0,0,0,0,0,0 |
| 9 | 0,0,0,0,0,0,0,0,1,0,0,0,0,0,0,0,0,0,0,0,0,0 |
| 10 | 0,0,0,0,0,0,0,0,0,1,0,0,0,0,0,0,0,0,0,0,0,0 |
| 11 | 0,0,0,0,0,0,0,0,0,0,1,0,0,0,0,0,0,0,0,0,0,0 |
| 12 | 0,0,0,0,0,0,0,0,0,0,0,1,0,0,0,0,0,0,0,0,0,0 |
| 13 | 0,0,0,0,0,0,0,0,0,0,0,0,1,0,0,0,0,0,0,0,0,0 |
| 14 | 0,0,0,0,0,0,0,0,0,0,0,0,0,1,0,0,0,0,0,0,0,0 |
| 15 | 0,0,0,0,0,0,0,0,0,0,0,0,0,0,1,0,0,0,0,0,0,0 |
| 16 | 0,0,0,0,0,0,0,0,0,0,0,0,0,0,0,1,0,0,0,0,0,0 |
| 17 | 0,0,0,0,0,0,0,0,0,0,0,0,0,0,0,0,1,0,0,0,0,0 |
| 18 | 0,0,0,0,0,0,0,0,0,0,0,0,0,0,0,0,0,1,0,0,0,0 |
| 19 | 0,0,0,0,0,0,0,0,0,0,0,0,0,0,0,0,0,0,1,0,0,0 |
| 20 | 0,0,0,0,0,0,0,0,0,0,0,0,0,0,0,0,0,0,0,1,0,0 |
| 21 | 0,0,0,0,0,0,0,0,0,0,0,0,0,0,0,0,0,0,0,0,1,0 |
| 22 | 0,0,0,0,0,0,0,0,0,0,0,0,0,0,0,0,0,0,0,0,0,1 |

*Table 3. Output codes for one-per-class approach.*

Table 3 shows the output codes (targets) learnt by the neural network when the input pattern $\vec{x} \in k$ user. We can observe that just the output number *k* is activated, and the number of outputs is equal to the number of users.

Table 4 shows the output codes (targets) where each user has his own code, and these codes are selected from the first 22 BCH (15,7) codes. In fact, BCH (15,7) yields up to $2^7 = 128$ output codes. However, we just need 22, because this is the number of users. On the other hand, we can see that the first two outputs are always the same. Thus, we can remove this outputs and our neural network will consist of just 13 outputs. It is interesting to observe that:

- BCH (15, 7) code (table 4) provides a minimum distance of 5 bits between different codes, while table 3 just provides a minimum distance of 2.
- BCH (15, 7) provides a more balanced amount of ones and zeros, while in table 3 almost all the outputs will be inhibitory.

| *k* | BCH (15,7) code |
|---|---|
| 1 | 0,0,0,0,0,0,0,0,0,0,0,0,0,0,0 |
| 2 | 0,0,0,0,0,0,1,1,1,0,1,0,0,0,1 |
| 3 | 0,0,0,0,0,1,0,0,1,1,1,0,0,1,1 |
| 4 | 0,0,0,0,0,1,1,1,0,1,0,0,0,1,0 |
| 5 | 0,0,0,0,1,0,0,1,1,1,0,0,1,1,0 |
| 6 | 0,0,0,0,1,0,1,0,0,1,1,0,1,1,1 |
| 7 | 0,0,0,0,1,1,0,1,0,0,1,0,1,0,1 |
| 8 | 0,0,0,0,1,1,1,0,1,0,0,0,1,0,0 |
| 9 | 0,0,0,1,0,0,0,0,0,0,1,1,1,0,1 |
| 10 | 0,0,0,1,0,0,1,1,1,0,0,1,1,0,0 |
| 11 | 0,0,0,1,0,1,0,0,1,1,0,1,1,1,0 |
| 12 | 0,0,0,1,0,1,1,1,0,1,1,1,1,1,1 |
| 13 | 0,0,0,1,1,0,0,1,1,1,1,1,0,1,1 |
| 14 | 0,0,0,1,1,0,1,0,0,1,0,1,0,1,0 |
| 15 | 0,0,0,1,1,1,0,1,0,0,0,1,0,0,0 |
| 16 | 0,0,0,1,1,1,0,1,0,0,0,1,0,0,0 |
| 17 | 0,0,0,1,1,1,1,0,1,0,1,1,0,0,1 |
| 18 | 0,0,1,0,0,0,0,0,0,1,1,1,0,1,0 |
| 19 | 0,0,1,0,0,0,1,1,1,1,0,1,0,1,1 |
| 20 | 0,0,1,0,0,1,0,0,1,0,0,1,0,0,1 |
| 21 | 0,0,1,0,0,1,1,1,0,0,1,1,0,0,0 |
| 22 | 0,0,1,0,1,0,0,1,1,0,1,1,1,0,0 |

*Table 4. 22 Output codes for BCH (15,7) approach.*

**Experimental Results**

We use a Multi-layer perceptron with 9 inputs, and *h* hidden neurons, both of them with *tansig* nonlinear transfer function (see figure 12).

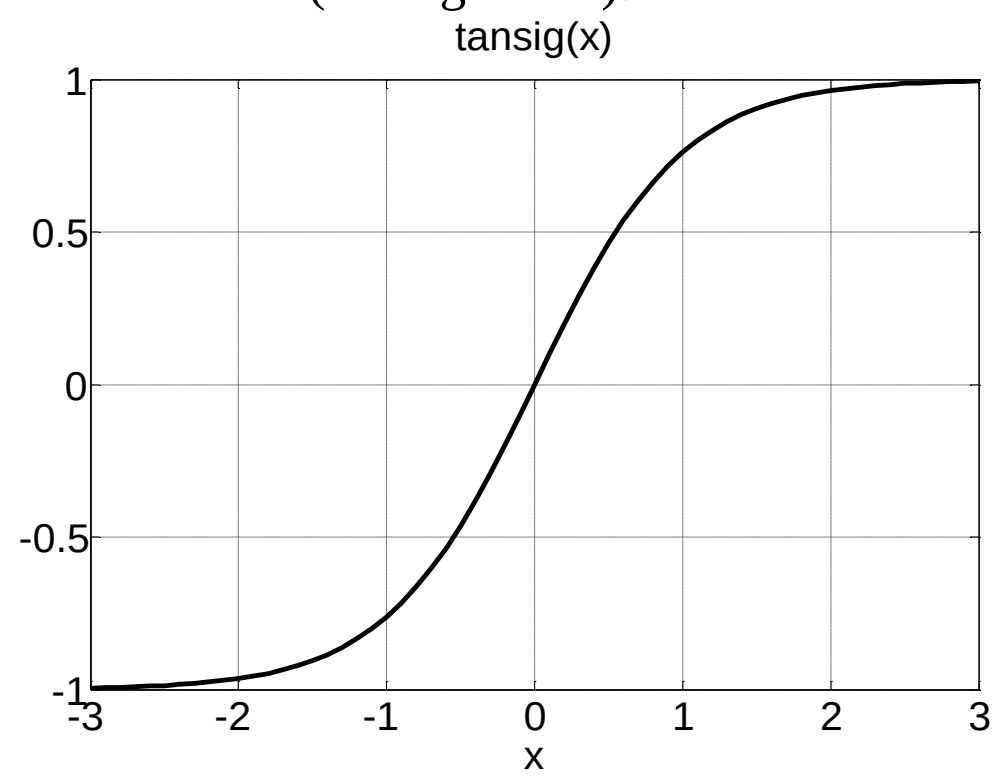

*Figure 12. Tansig non-linear transfer function.*

This function is symmetrical around the origin. Thus, we modify the output codes shown in tables 3 & 4 replacing each "0" by "–1". In addition, we normalize the input vectors $\vec{x}$ for zero mean and maximum modulus equal to 1.

Figure 13 shows the histograms of the neural net outputs for genuine scores (top) and impostors (bottom). A fitted Gaussian is also plotted for each distribution. Figure 14 and 15 show the same information when using BCH (15,7) output codes during training (targets). Figure 14 corresponds to MSE and figure 15 to MAD (see equations 4 and 5) computation between expected values (shown in table 4, changing "0" per "–1") and obtained outputs. Obviously equations 4 and 5 yield a resulting *distance*, which is always greater or equal to zero. For this motivation, and for comparison purposes, we have plotted (1 – *distance*). For this reason, the maximum value in figures 14 and 15 is equal to 1.

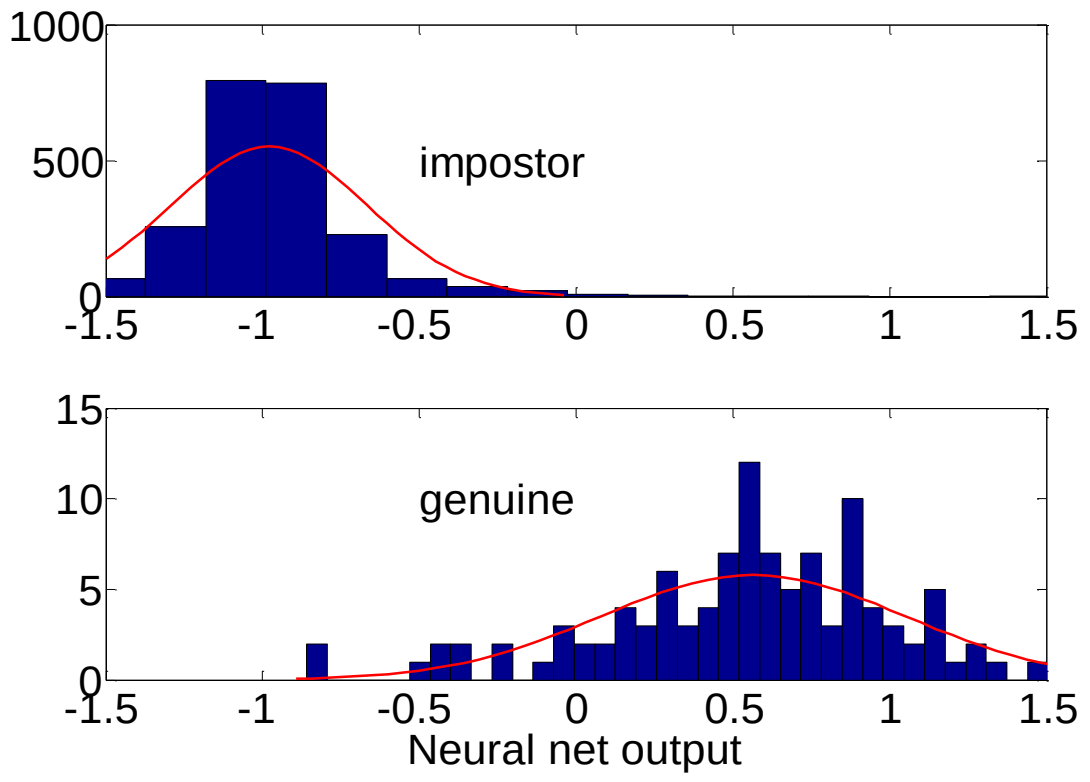

*Figure 13. Neural net output histograms for one-per-class approach.*

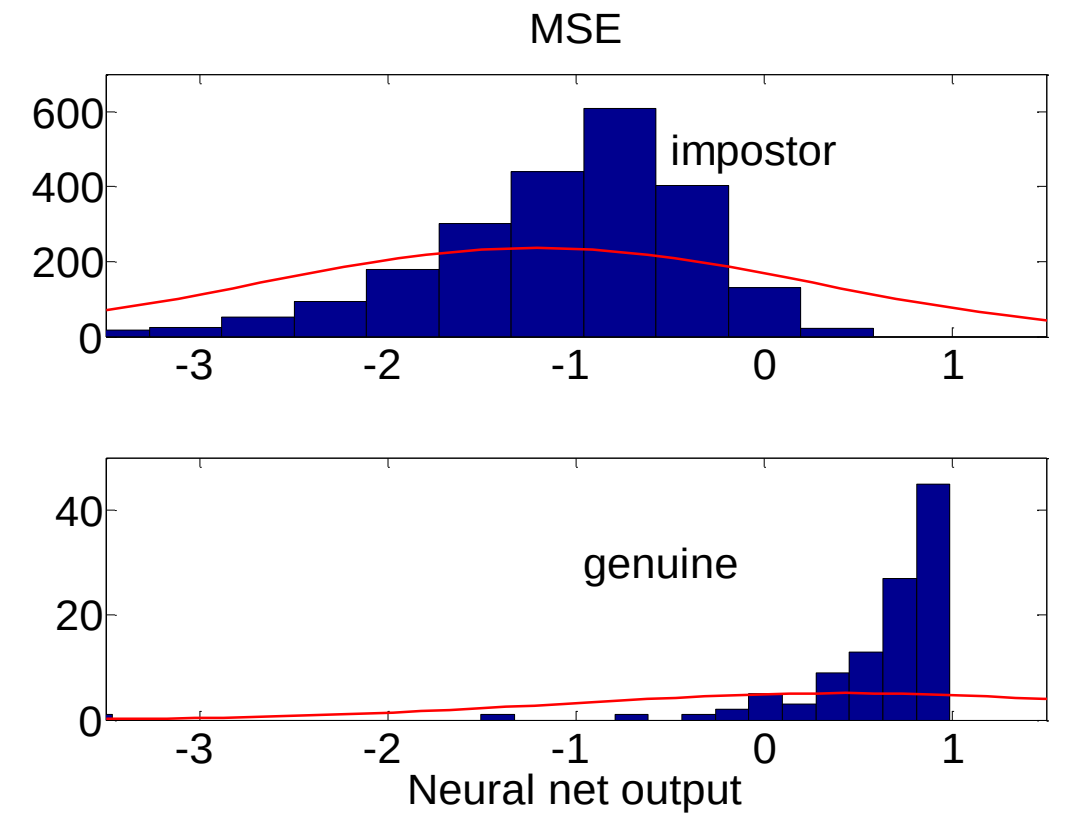

*Figure 14. MLP output histograms for error-correcting output coding approach, using MSE computation.*

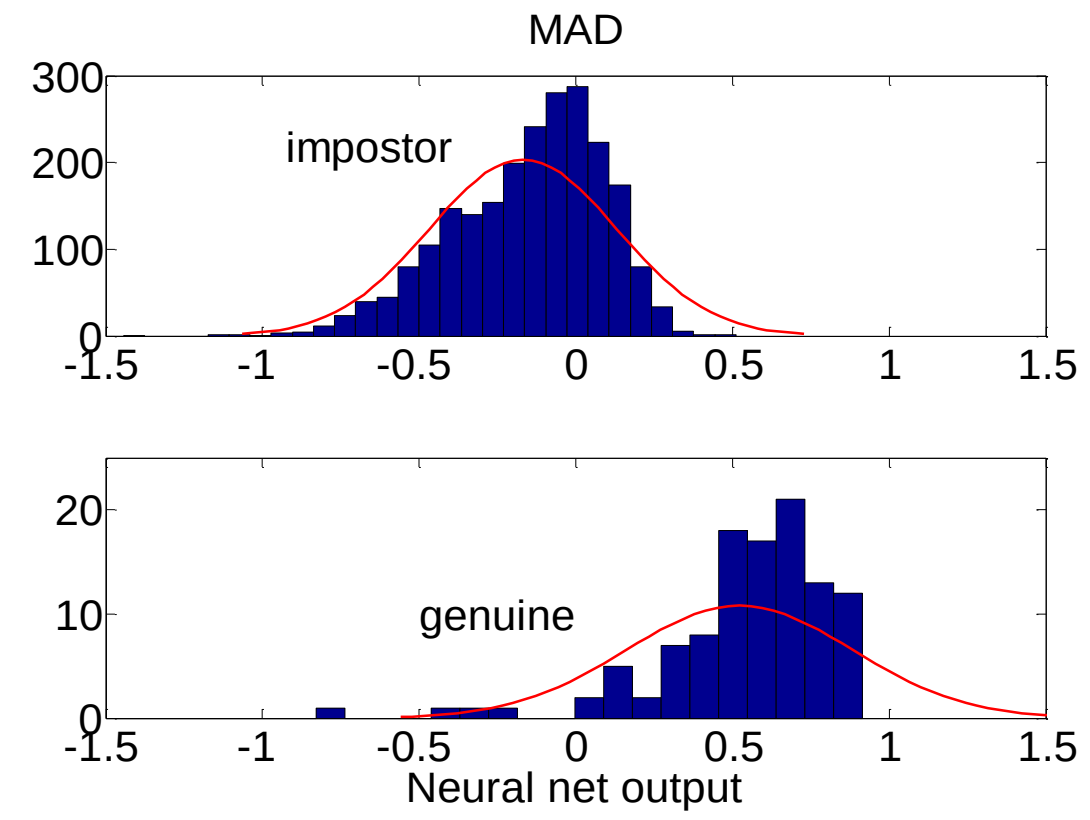

*Figure 15. MLP output histograms for error-correcting output coding approach, using MAD computation.*

We will summarize the Multi-Layer Perceptron number of neurons in each layer using the following nomenclature: *inputs*× *hidden*× *output*. In our experiments, the number of inputs is fixed to 9, and the other parameters can vary according to the selected strategy.

Table 5 shows the experimental results using one neural net per person with one output (total set: 22 MLP). Tables 6 and 7 show the results of a single MLP with 22 outputs (trained using the target shown in table 3). Table 6 has been obtained with 30 hidden neurons, and table 7 with 40 hidden neurons. Table 8 shows the results using BCH (31, 6), table 9 with BCH (15, 7), and table 10 with BCH (15,5).

| 22 MLP epoch | Identification rate (%) | | | Min(DCF) (%) | | |
|---|---|---|---|---|---|---|
| | mean | σ | max | mean | σ | min |
| 10 | 84.12 | 3.08 | 91.82 | 5.38 | 0.91 | 3.2 |
| 50 | 84.37 | 2.99 | 91.82 | 5.27 | 0.71 | 3.4 |

Table 5. 22 MLP 9×30×1 (1 output per MLP).

| 1 MLP epoch | Identification rate (%) | | | Min(DCF) (%) | | |
|---|---|---|---|---|---|---|
| | mean | σ | max | mean | σ | min |
| 10 | 87.55 | 2.36 | 91,82 | 4.95 | 0.82 | 2.92 |
| 50 | 88.02 | 1.91 | 92,73 | 4.79 | 0.75 | 2.94 |

Table 6. 1 MLP 9×30×22 (one-per-class).

| Epoch | Identif. rate (%) | | | Min(DCF) (%) | | |
|---|---|---|---|---|---|---|
| | mean | σ | max | mean | σ | min |
| 10 | 88.23 | 2.24 | 92.73 | 4.70 | 0.76 | 3.03 |
| 50 | 88.01 | 1.88 | 91.82 | 4.81 | 0.73 | 3.03 |

Table 7. 1 MLP 9×40×22 (one-per-class).

| | Epoch | Identif. rate (%) | | | Min(DCF) (%) | | |
|---|---|---|---|---|---|---|---|
| | | mean | σ | max | mean | σ | min |
| MAD | 10 | 88.75 | 1.75 | 92.73 | 5.57 | 0.79 | 3.53 |
| | 50 | 88.46 | 2.01 | 93.64 | 5.71 | 0.89 | 3.53 |
| MSE | 10 | 88.91 | 1.81 | 92.73 | 5.63 | 0.81 | 3.57 |
| | 50 | 88.56 | 2.05 | 92.73 | 5.75 | 0.93 | 3.40 |

Table 8. 1 MLP 9×40×30 (ECOC BCH (31, 6)).

| | Epoch | Identif. rate (%) | | | Min(DCF) (%) | | |
|---|---|---|---|---|---|---|---|
| | | mean | σ | max | mean | σ | min |
| MAD | 10 | 84.79 | 2.27 | 90 | 7.35 | 0.82 | 5.82 |
| | 50 | 85.23 | 2.29 | 90 | 7.37 | 0.94 | 4.98 |
| MSE | 10 | 85.07 | 2.27 | 90 | 6.73 | 0.83 | 4.72 |
| | 50 | 85.42 | 2.30 | 90 | 6.72 | 0.84 | 4.94 |

Table 9. 1 MLP 9×40×13 (ECOC BCH (15, 7)).

| | Epoch | Identif. rate (%) | | | Min(DCF) (%) | | |
|---|---|---|---|---|---|---|---|
| | | mean | σ | max | mean | σ | min |
| MAD | 10 | 87.86 | 1.75 | 91.82 | 5.99 | 0.77 | 3.98 |
| | 50 | 87.34 | 1.77 | 90.91 | 6.40 | 0.95 | 4.37 |
| MSE | 10 | 88.03 | 1.58 | 90.91 | 5.83 | 0.74 | 4.05 |
| | 50 | 87.34 | 1.89 | 90.91 | 6.21 | 1.0 | 3.85 |

Table 10. 1 MLP 9×40×15 (ECOC BCH (15, 5)).

A good error-correcting output code for a *k*-class problem should satisfy two properties [13]:

- Row separation: each codeword should be well-separated in Hamming distance from each of the other codewords.
- Column separation: each bit-position function $f_i$ should be uncorrelated from the functions to be learnt for the other bit positions $f_j$, $j \neq i$.

Error- correcting codes only succeed if the errors made in the individual bit positions are relatively uncorrelated,

so that the number of simultaneous errors in many bit positions is small. For this purpose, we have used the algorithm proposed in [18] for random ECOC generation. Table 11 summarizes the obtained results.

For 30 outputs (classifiers) and 22 users we get a minimum row (class) Hamming distance $H_c$ = 18 bit and minimum column (classifier) Hamming distance $H_L$ = 12 bit, after 500 random iterations. For 15 outputs we get $H_c$ = 8 bit and $H_L$ = 14 bit, after 500 random iterations.

| | Epoch | Identif. rate (%) | | | Min(DCF) (%) | | |
|---|---|---|---|---|---|---|---|
| | | mean | σ | max | mean | σ | min |
| MAD | 10 | 86.26 | 2.58 | 90.91 | 7.53 | 0.93 | 5.26 |
| | 50 | 86.2 | 2.39 | 92.73 | 7.72 | 1.05 | 5.19 |
| MSE | 10 | 86.46 | 2.53 | 90.91 | 6.62 | 0.87 | 4.63 |
| | 50 | 86.27 | 2.50 | 92.73 | 6.81 | 1.09 | 4.46 |

Table 11. 1 MLP 9×40×30 (random ECOC generation).

| | Epoch | Identif. rate (%) | | | Min(DCF) (%) | | |
|---|---|---|---|---|---|---|---|
| | | mean | σ | max | mean | σ | min |
| MAD | 10 | 85.02 | 2.31 | 90.91 | 8.54 | 1.05 | 5.58 |
| | 50 | 85.05 | 2.50 | 90 | 8.73 | 1.07 | 5.82 |
| MSE | 10 | 85.19 | 2.34 | 90.91 | 6.94 | 0.99 | 3.98 |
| | 50 | 85.17 | 2.41 | 90 | 7.05 | 1.03 | 4.59 |

Table 12. 1 MLP 9×40×15 (random ECOC generation).

## 5. CONCLUSIONS

Taking into account the experimental results, we observe the following conclusions:

- MLP classifiers outperform the classical Nearest Neighbor classifier.
- Comparing tables 5 and 6, we observe that better results are obtained using a single MLP with as many outputs as users. However, if the number of users were larger, a single network can be hard to train. However, we have made 100 random initializations for 22 MLP, and we have worked out the statistics for all these experiments. Another possibility would be to pick up, for each user, the best random initialization and then to provide statistical results for the 22 selected MLP. This latter experiment would imply a third partition of the database (one for training, one for MLP selection, and another one for statistical computations). We have not performed it due to the lack of enough testing samples.
- Comparing table 9 versus table 10 we see better performance when using BCH (15, 5). This is in agreement with the fact that this code has higher correction capability than BCH (15, 7).
- Comparing table 8 versus table 10 we see better performance when using BCH (31, 6). This is in agreement with the fact that this code has higher correction capability than BCH (15, 5). However, we must take into account that 30 outputs imply much more weights than 15 outputs.
- Although it is supposed that random generation for ECOC (maximizing row and column minimum distances we get $H_c$ = 18 bit and $H_L$ = 12 bit for 30 outputs) should outperform BCH codes ($H_c$ = 14 bit, $H_L$ = 20 bit for BCH (15, 5) 15 outputs, and $H_c$ = 10 bit, $H_L$ = 0 bit for BCH (15, 7) 13 outputs), our experimental results reveal better performance when using the latest ones. On the other hand, it is a slight improvement over one-per-output approach, and just for identification, not for verification application.